\documentclass[conference]{IEEEtran}
\IEEEoverridecommandlockouts

\usepackage{cite}
\usepackage{amsmath,amssymb,amsfonts}
\usepackage{algorithmic}
\usepackage{graphicx}
\usepackage{textcomp}
\usepackage{xcolor}
\def\BibTeX{{\rm B\kern-.05em{\sc i\kern-.025em b}\kern-.08em
    T\kern-.1667em\lower.7ex\hbox{E}\kern-.125emX}}

\usepackage{orcidlink}
\usepackage{hyperref}
\usepackage{verbatim}
\usepackage{threeparttable}
\usepackage{xspace,hyphenat}

\usepackage[hide]{mnotes}             

\Mnewauthor{SL}{orange}

\begin{document}

\title{Predicting Psychological Well-Being from Spontaneous Speech using LLMs\\
}

\author{

\IEEEauthorblockN{
Erfan~Loweimi$^{1,2\dagger}$~\orcidlink{0000-0002-8761-021X}
\thanks{$^{\dagger}$This work was conducted while the author was with the Centre for Medical Informatics (CMI), Usher Institute, University of Edinburgh.}
}
\IEEEauthorblockA{$^{1}$\textit{Cisco}, UK}
\IEEEauthorblockA{$^{2}$\textit{University of Edinburgh}, Edinburgh, UK \\
E.Loweimi@ed.ac.uk}

\and

\IEEEauthorblockN{
Sofia~de~la~Fuente~Garcia~\orcidlink{0000-0002-4915-3177}
}
\IEEEauthorblockA{\textit{Centre for Medical Informatics (CMI)} \\
\textit{Usher Institute} \\
\textit{University of Edinburgh}, Edinburgh, UK \\
Sofia.Delafuente@ed.ac.uk}

\and

\IEEEauthorblockN{
Saturnino~Luz~\orcidlink{0000-0001-8430-7875}
}
\IEEEauthorblockA{\textit{Centre for Medical Informatics (CMI)} \\
\textit{Usher Institute} \\
\textit{University of Edinburgh}, Edinburgh, UK \\
S.Luz@ed.ac.uk}

}

\maketitle

\begin{abstract}
  We investigate the use of Large Language Models (LLMs) for zero-shot
  prediction of Ryff Psychological Well-Being (PWB) scores from
  spontaneous speech. Using a few minutes of voice recordings from 111
  participants in the PsyVoiD database, we evaluated 12 instruction-tuned
  LLMs, including Llama-3 (8B, 70B), Ministral, Mistral, Gemma-2-9B,
  Gemma-3 (1B, 4B, 27B), Phi-4, DeepSeek (Qwen and Llama), and
  QwQ-Preview. A domain-informed prompt was developed in collaboration
  with experts in clinical psychology and linguistics. Results show
  that LLMs can extract semantically meaningful cues from spontaneous
  speech, achieving Spearman correlations of up to 0.8 on 80\% of the
  data. Additionally, to enhance explainability, we conducted
  statistical analyses to characterise prediction variability and
  systematic biases, alongside keyword-based word cloud analyses to
  highlight the linguistic features driving the models' predictions.
\end{abstract}


\section{Introduction}
\label{sec:intro}

Psychological well-being is central to overall health, resilience to adverse events, and daily functioning. Recent global trends highlight its importance, with mental health challenges affecting millions annually \cite{salari2020prevalence}. For example, the COVID-19 pandemic and its wide-ranging effects on health, economy, and other domains substantially intensified psychological distress worldwide \cite{world2022mental,le2022psychological}. Timely identification and longitudinal monitoring of well-being are therefore essential.

Conventional assessment of psychological well-being relies on clinical interviews \cite{von1987anxiety,stade2023depression} and self-report questionnaires. Although informative, these methods are subjective, resource-intensive, and face significant scalability challenges \cite{newson2020heterogeneity,nordgaard2013psychiatric}. Because spoken language inherently encodes internal states through both acoustic and linguistic cues \cite{wu2011automatic}, recent advances in AI have created new opportunities to harness these signals for non-invasive, low-cost screening \cite{delafuenteritchieluz2020JADreview,roy2025llmmental,low2020automated,busch2025llm,wagner2025scalable}.

Large language models (LLMs) enhance speech-based assessment by extracting psychological markers from spontaneous language. They have been shown to approximate clinical scoring systems such as the Hospital Anxiety and Depression Scale (HADS) \cite{zigmond1983hospital}, achieving reasonable agreement with human-coded assessments \cite{loweimi2025interspeechHADS}. Similar approaches have been applied to depression detection from speech using multimodal LLM architectures \cite{li2025depression,patapati2024trimodal}. With carefully designed prompts, LLMs can serve as rapid and scalable proxies for traditional screening tools, especially where annotated data are limited or costly to obtain. Despite this promise in mental health monitoring \cite{guo2024large,hua2025scoping}, the prediction of psychological well-being from spontaneous speech in \emph{zero-shot} \cite{zeroshot-2009,llm-few-shot} settings, namely without task-specific fine-tuning, remains underexplored.

Building on this, we extend the zero-shot evaluation paradigm from predicting symptom-focused measures such as HADS to Ryff’s Psychological Well-Being (PWB) framework \cite{ryff1989happiness,ryff1995structure}. 
Unlike clinical tools that primarily assess distress or dysfunction, Ryff’s framework provides a holistic, eudaimonic account of well-being, making it a rigorous testbed for evaluating whether LLMs can infer higher-order psychological constructs from spontaneous speech \cite{ryff2014self}.

Recent work has questioned whether human-centric frameworks like Ryff’s align with how LLMs conceptualise well-being. For instance, \cite{lau2025human} analysed LLM responses to open-ended prompts on ``flourishing'' and introduced the \textit{PAPERS} framework (Purposeful Contribution, Adaptive Growth, Positive Relationality, Ethical Integrity, Robust Functionality, Self-Actualised Autonomy). Their findings suggest that LLMs generate internally coherent but machine-oriented accounts of well-being, emphasising effectiveness and compliance with instructions over autonomy or existential meaning. This raises a central question: do LLMs capture genuine markers of human psychological states, or merely computational analogues that approximate them?

To address this question, we predict Ryff’s scales to evaluate how well LLMs can infer PWB from unstructured personal narratives. Specifically, we test whether instruction-tuned LLMs can estimate Ryff PWB scores \cite{ryff1989happiness,ryff1995structure} from short spontaneous speech recordings collected during the COVID-19 lockdown as part of the PsyVoiD dataset \cite{FuenteLuz@2023psyvoidDataShare}. Participants provided brief monologues describing their daily experiences under lockdown. Performance is reported using both Pearson correlation coefficient (PCC) and Spearman correlation coefficients (SCC), complemented by additional statistical tests.


Our contributions are threefold:
\begin{itemize}
    \item \textbf{Zero-shot well-being prediction:} We evaluate twelve instruction-tuned large language models (LLMs)---Meta-Llama \cite{meta-llama} (3.1-8B \cite{llama3-8b} and 3.3-70B \cite{meta-llama}), Microsoft Phi-4 \cite{phi-4}, Google Gemma-2-9B \cite{gemma-2}, Google Gemma-3 (1B, 4B, 27B) \cite{gemma-3}, Ministral-2410 \cite{ministral}, Mistral-NeMo-2407 \cite{mistral2024nemo}, QwQ-32B-Preview \cite{qwq-32b-preview,qwen2}, DeepSeek-R1-Distill-Qwen-32B (DeepSeek Qwen) \cite{deepseek-qwen}, and DeepSeek-R1-Distill-Llama-70B (DeepSeek Llama) \cite{deepseek-qwen}---for zero-shot prediction of Ryff’s Psychological Well-Being (PWB) dimensions from spontaneous speech transcripts.
    \item \textbf{Psychologically informed prompt design:} We develop and evaluate prompts that integrate established prompt engineering strategies with domain knowledge from psychological well-being research to guide LLM reasoning and output structure.
    \item \textbf{Model behaviour analysis and interpretability:} We conduct extensive statistical analyses and linguistic profiling of LLM outputs to characterise behavioural patterns and linguistic cues associated with well-being prediction.
\end{itemize}

The rest of this paper is structured as follows. After describing the data and the Ryff scale in Section~\ref{sec:method}, Section~\ref{sec:workflow} presents the workflow, including the prompt engineering approach (Section~\ref{sec:prompt}); Section~\ref{sec:exp} presents results along with discussion, statistical analysis, and keyword visualisation; Section~\ref{sec:conclusion} concludes the paper.

\section{Psychological Assessment}
\label{sec:method}

\subsection{PsyVoiD dataset}
The PsyVoiD dataset 
\cite{FuenteLuz@2023psyvoidDataShare} was collected through a large-scale, anonymous survey and comprises 111 participants (70 female, 41 male), aged 21--86, all residing in Scotland during the COVID-19 lockdown. Of these, 34 participants (31\%) reported a history of depression. Each recording lasts one to two minutes, containing an average of 150 words and 92 unique words per sample, with an average articulation rate of approximately 2 words per second. Table~\ref{tab:psyvoid-stats} presents descriptive statistics (mean, median, standard deviation (STD), minimum, and maximum) for some dataset attributes.

\subsection{Psychological Well-being Measurement}
The reference measure for psychological assessment in this study is the Ryff Psychological Well-Being (PWB) scales \cite{ryff1989happiness,ryff1995structure}, a widely validated self-report instrument. The PWB framework comprises six dimensions: \textit{autonomy}, \textit{environmental mastery}, \textit{personal growth}, \textit{positive relations with others}, \textit{purpose in life}, and \textit{self-acceptance}. Items are rated on a Likert-type scale, with higher values reflecting greater well-being (with reverse-keyed items scored accordingly). Subscale scores are obtained by aggregating item responses for each dimension; an overall index can also be derived following standard scoring practice. Descriptive statistics for the Ryff PWB scale are also reported in Table~\ref{tab:psyvoid-stats}.
\mnSL{It may be interesting, perhaps in a future paper, to compare Ryff and HADS prediction; one could start with assessing the correlation between the two.}

\begin{table}[t]
    \centering
    \begin{threeparttable}
        \caption{Descriptive statistics on PsyVoiD's 111 subjects}
        \label{tab:psyvoid-stats}
        \small
        \begin{tabular}{l|ccccc}
            \hline 
            Variable        & Mean & Median  & STD & Min & Max \\ 
            \hline
            Duration (sec)  & 72.2 & 76.9 & 27.2 & 14.6 & 117.4 \\ 
            Age       & 59   & 62   & 12.9 & 21   &  86   \\
            \hline
            \#Words         &152   &158   & 77   &  9   & 310   \\ 
            \#UniqueWords   & 92   & 97   & 38   &  9   & 169   \\
            \#WordsPerSec   & 2.1  &  2.0 &  0.5 &  0.4 &   3.2 \\ 
            \hline
            Ryff PWB        & 91.7 &  95.0 &  16.8 &  51 & 123 \\  
            \hline 
        \end{tabular}
    \end{threeparttable}
\end{table}

\begin{figure}[t]
  \centering
  \includegraphics[width=\linewidth]{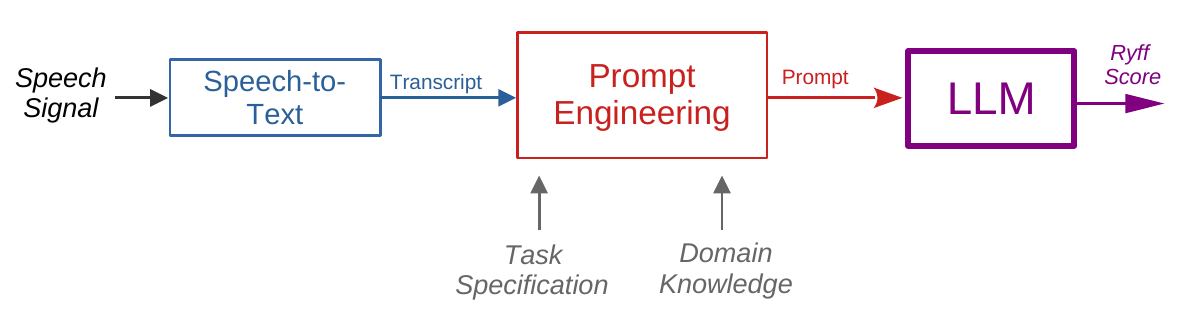}
  \caption{Workflow for zero-shot Ryff well-being estimation: ASR front-end, prompt stage, and LLM inference.}
  \label{fig:workflow}
\end{figure}

\begin{table}[h!]
    \centering
    \begin{threeparttable}
        \caption{WER on PsyVoiD for various Whisper models}
        \label{tab:wer}
        \small
        \begin{tabular}{l|c|cccc}
            \hline 
            Model    & \#Params (B)    & WER  & Sub & Del & Ins \\ 
            \hline
            Large (v3)      & 1.550 &  9.2 & 3.0 & 4.0 & 2.2 \\  
            Medium (en)     &  0.769 & 10.0 & 3.3 & 4.8 & 1.9 \\
            Small (en)      &  0.244 & 13.1 & 3.8 & 4.6 & 4.7 \\
            \hline 
        \end{tabular}
    \end{threeparttable}
\end{table}

\section{Workflow}
\label{sec:workflow}

Fig.~\ref{fig:workflow} presents the system architecture, comprising a speech-to-text (automatic speech recognition) front-end, prompt engineering module, and LLM-based decision back-end.

\subsection{Speech-to-Text Conversion}
Speech recordings can be converted into text either through manual annotation or by using automatic speech recognition (ASR) systems. Recent state-of-the-art ASR models, such as OpenAI’s Whisper \cite{whisper-2022}, achieve strong performance and are robust against noise, speaker variability, and spontaneous speech. Nevertheless, transcription errors remain non-negligible.

As shown in Table~\ref{tab:wer}, even Whisper \textit{Large-v3} exhibits a Word Error Rate (WER) of approximately 9.2\% on the PsyVoiD data, meaning that on average, one out of every eleven words is transcribed incorrectly. In addition to typical substitution, deletion, and insertion errors, ASR models can also produce hallucinations, such as generating repetitive or extraneous phrases that are not present in the original speech. These transcription errors, including both distortions of linguistic content and hallucinations, can mislead the language model and compromise the analysis of psychologically relevant features such as hesitation markers, self-referential language, and affective expressions.

To remove the confounding effect of ASR errors and ensure that the input text accurately reflects the original speech, we therefore rely on manually annotated transcripts in this study. This choice allows us to isolate the performance of the downstream large language models and ensures that the observed effects are attributable to linguistic modelling rather than transcription noise.
\mnSL{Another possible extension would be comparing performance on manual vs ASR transcripts.}

\subsection{LLM Prompt Engineering}
\label{sec:prompt}
We feed the manual transcripts to LLMs to estimate psychological well-being (i.e., Ryff scores) from spontaneous speech. In contrast to conventional supervised approaches, which require task-specific training, LLMs can operate in a \emph{zero-shot} regime \cite{zeroshot-2009,llm-few-shot}, leveraging broad linguistic and psychosocial priors learnt during pre-training to infer relevant constructs \cite{guo2024large}. To enhance reliability and interpretability, we design a domain-informed prompt co-developed with input from clinical psychology and linguistics, framing the task as a structured assessment conducted by a clinician--linguist team. This role-based prompting strategy has been shown to better align model outputs with expert reasoning in domain-specific tasks \cite{guo2024large,priyadarshana2024prompt}. 

\begin{figure}[t]
  \centering
  \includegraphics[width=\linewidth]{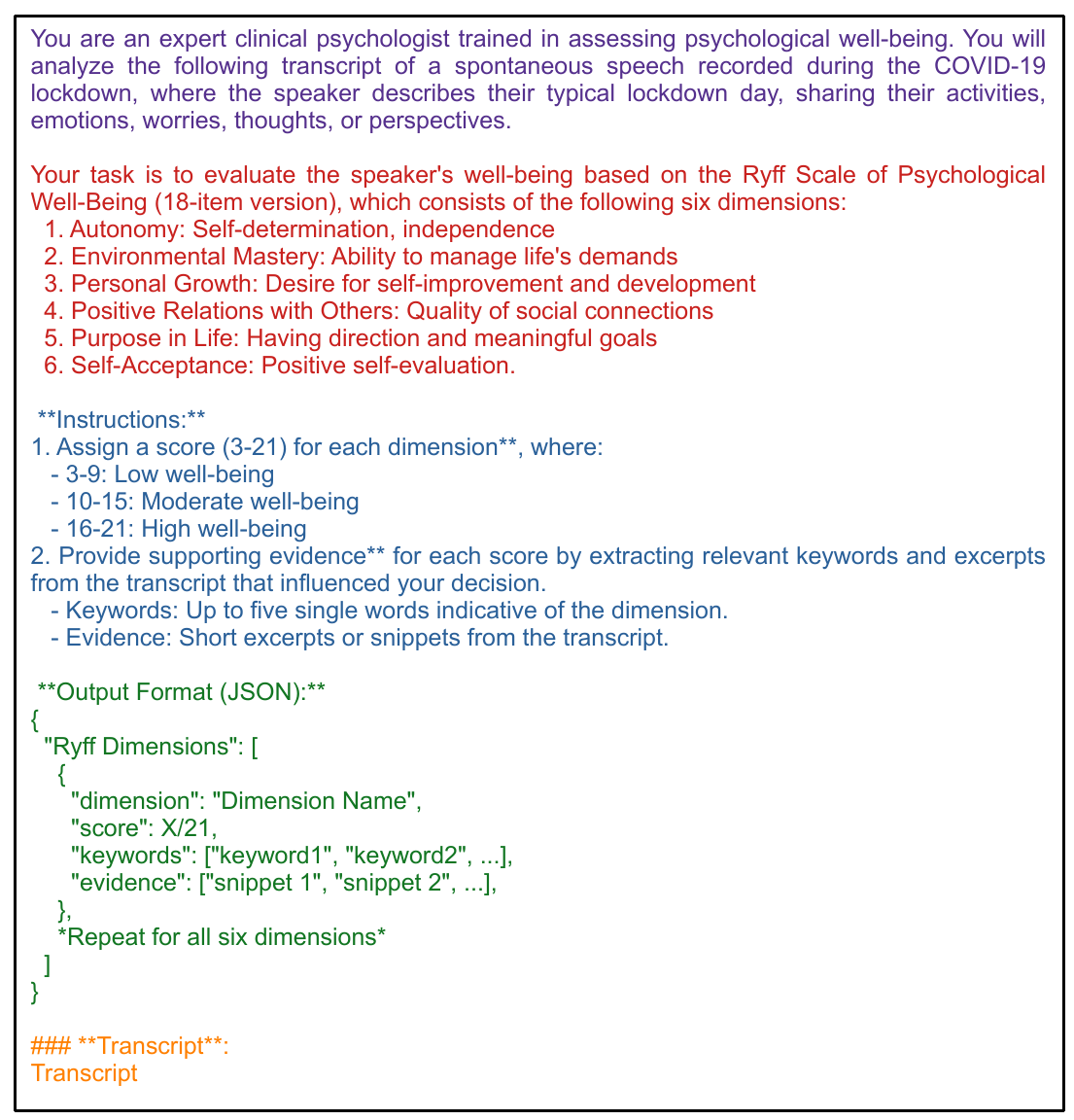}
  \caption{Prompt design for Ryff PWB inference via LLMs: each colour corresponds to a different aspect of the prompt.}
  \label{fig:prompt}
\end{figure}

The prompt (Fig.~\ref{fig:prompt}) is designed to guide the LLM as an expert clinical psychologist evaluating psychological well-being from spontaneous speech. The model analyses transcripts of participants describing their typical day during the COVID-19 lockdown, mapping content to the six Ryff PWB dimensions: Autonomy, Environmental Mastery, Personal Growth, Positive Relations with Others, Purpose in Life, and Self-Acceptance. For each dimension, the LLM assigns a score (3--21), interprets the meaning of low, moderate, and high well-being, and provides supporting evidence by extracting indicative keywords and relevant transcript excerpts. The predicted overall Ryff score is calculated as the sum of all six dimension scores.
The output is structured in JSON format, with scores, keywords, and evidence for all six dimensions. 

By combining this psychologically grounded role-based prompting with structured linguistic analysis and explicit justifications, the approach enhances robustness, explainability, and transparent interpretability in zero-shot LLM-based assessments of multidimensional well-being.

\section{Experimental Results and Discussion}
\label{sec:exp}

\subsection{Performance Evaluation}
Table~\ref{tab:cc} reports the Pearson and Spearman correlations between the ground-truth Ryff scores, derived from standard questionnaires completed by participants, and the Ryff scores predicted by the LLMs. The highest Pearson correlation coefficients are achieved by Meta-Llama-3.3-70B and DeepSeek-Qwen, while the highest Spearman correlations are observed for DeepSeek-Llama and Meta-Llama-3.3-70B.\footnote{As a practical note, in our experiments, the 
DeepSeek-R1-Distill-Llama-70B and Llama-3.3-70B-Instruct models---both very large models with 70B parameters---were quantised to 8-bit precision due to GPU memory constraints, while all other models were run using bfloat16 precision. Consequently, the reported results for these two models may not fully reflect their performance under higher-precision settings.}

Although model rankings are broadly similar across both correlation metrics, some differences highlight the need to consider which metric is more appropriate. Pearson correlation assumes a linear relationship and Gaussian-distributed data; however, these assumptions are violated for Ryff scores, which are bounded and skewed (as shown later in Fig.~\ref{fig:hist-ryff}). In contrast, Spearman correlation, being rank-based, is robust to these issues, providing a more reliable measure of agreement between predicted and actual Ryff scores in this context.

Are all of these correlations statistically significant? To answer this, we computed p-values for both Pearson and Spearman correlations using the \texttt{scipy.stats} library \cite{pedregosa2011scikit}. Specifically, \texttt{scipy.stats.pearsonr} returns the Pearson correlation coefficient ($r$) along with a two-tailed p-value under the null hypothesis $H_0: r = 0$, assuming the data are drawn from a bivariate normal distribution. Similarly, \texttt{scipy.stats.spearmanr} computes the SCC ($\rho$) and a corresponding two-tailed p-value under $H_0: \rho = 0$. 
\mnSL{Another interesting question might be: Do the models correlate better among themselves than with the gold standard? BTW, have you set up a gitlab repo for this experiment, so we can play with the data and this sort of hypothesis?}

As shown in Table~\ref{tab:cc}, the p-values for most models are well below 0.01, indicating statistically significant correlations. Exceptions include Gemma-3-1B, for which the Pearson and Spearman p-values are 0.0285 and 0.0119, respectively; although higher, these values remain below the 0.05 threshold. In contrast, for DeepSeek-Llama---which achieved the highest SCC---the p-values are very high (0.885 for Pearson, 0.667 for Spearman), indicating that these correlations are not statistically significant.%
\mnSL{This looks a bit strange, given that all other comparisons yielded significant values, and the sample size is presumably the same across all comparisons. Perhaps plotting a scatterplot might shed some light on this. Also, it may be a good idea to correct these p-values for multiple testing, if not done already.}
While this occurs for only one LLM, it highlights the importance of reporting p-values alongside correlation coefficients in healthcare applications. High correlation coefficients alone can be misleading if sample size, variance, or distributional assumptions prevent statistical significance. Therefore, all results in Table~\ref{tab:cc} should be interpreted together with their p-values to ensure conclusions are statistically meaningful.

\subsection{Cumulative Correlation Analysis}
\begin{figure}[t]
  \centering
  \includegraphics[width=0.9\linewidth]{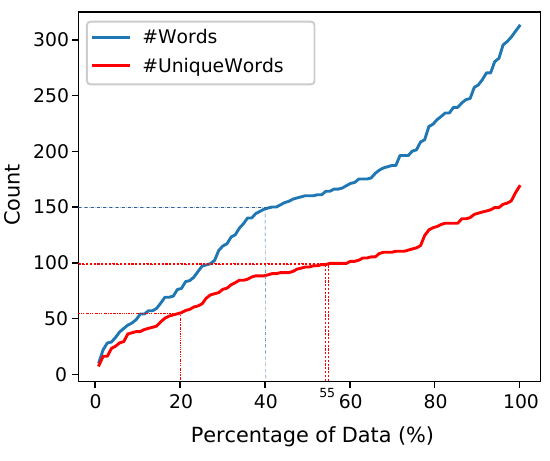}
  \caption{Word and unique word counts for different recordings in the PsyVoiD dataset. 40\% of the recordings contain fewer than 100 words, and 55\% contain fewer than 100 unique words.}
  \label{fig:word-count}
\end{figure}

The best correlation, 0.444 for Meta-Llama-3.3-70B, is statistically significant but modest, raising the question of whether limitations stem from the data or the LLM itself. It is important to note that estimating Ryff scores from only a few minutes of spontaneous speech is inherently difficult. As shown in Fig.~\ref{fig:word-count}, roughly 20\% of transcripts contain fewer than 60 unique words. While word count is only a coarse proxy for informational richness, it illustrates the scarcity of linguistic cues in some recordings. The overall Spearman correlation therefore reflects not only model performance but also these cases, where limited linguistic and contextual information constrains prediction accuracy. Such limitations pose challenges even for human experts, not just for the LLMs.

To further assess model reliability, we analysed Spearman correlations
under \textit{progressive data retention}. Files were sorted by the
absolute difference between LLM-predicted and ground-truth Ryff
scores, and cumulative correlations were computed iteratively as more
data were included ($n = 2 \dots N$, with $N$ as the total number of
files). Data retention is defined as in
Equation~(\ref{eq:dataretention}), reaching 100\% when all files are
included.

\begin{equation}
  \label{eq:dataretention}
  \mathrm{Data\;Retention \ (\%)} = \frac{n}{N} \times 100
\end{equation}

\begin{figure}[t]
  \centering
  \includegraphics[width=\linewidth]{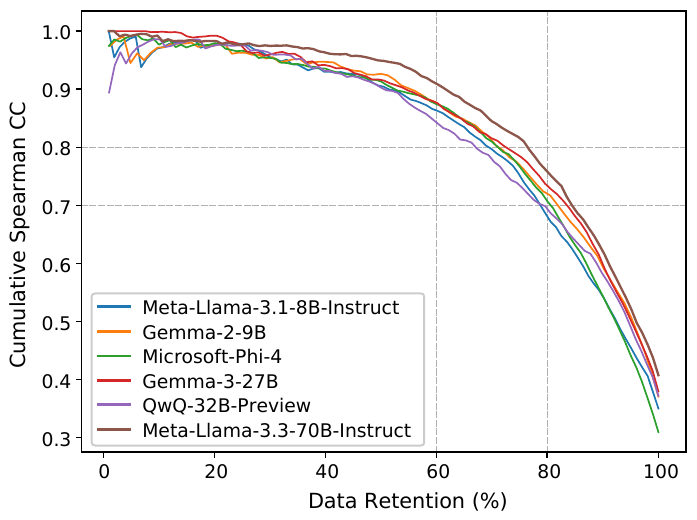}
  \caption{Cumulative analysis of the Spearman correlation coefficient (SCC) for various LLMs. For 80\% of the data, the SCC exceeds 0.7.}
  \label{fig:cum-scc}
\end{figure}

\mnSL{Set data retention formula out in eq environment as we
  have plenty of space.}
Fig.~\ref{fig:cum-scc} shows cumulative correlations as a function of data retention. As lower-quality samples are added, correlation gradually declines, reaching the full retention values reported in Table~\ref{tab:cc}. Notably, when considering only the top 75\% of samples, the Spearman correlation for the best model (Meta-Llama-3.3-70B) rises to 0.8, demonstrating the strong potential of LLMs for psychological inference when sufficient information is available.

\begin{table}[t]
    \centering
    \begin{threeparttable}
        \caption{Correlation between LLM predictions and ground-truth scores. \#Param is in billions, and PV denotes the p-value.}
        \label{tab:cc}
        \small
        \begin{tabular}{l|c|c|c|c|c}
        \hline
         LLM Name & \#Param  & PCC & SCC & $\text{PV}_{\text{PCC}}
$& $\text{PV}_{\text{SCC}}
$ \\  
        \hline
            Meta-Llama-3.1    & 8.0  & 0.359 & 0.350 & $<$0.01  & $<$0.01 \\  
            Ministral    & 8.0  & 0.330 & 0.317 & $<$0.01  & $<$0.01 \\  
            Mistral-NeMo & 12.2 & 0.401 & 0.388 &$<$0.01  & $<$0.01  \\  
            Microsoft-Phi-4   & 14.7 & 0.338 & 0.355 & $<$0.01  & $<$0.01  \\  
            QwQ-Preview       & 32.8 & 0.390 & 0.372 & $<$0.01  & $<$0.01  \\  
            DeepSeek-Qwen     & 32.8 & \bf 0.433 & 0.386 &$<$0.01 & $<$0.01 \\  
            DeepSeek-Llama    & 70.6 & 0.180 & \bf 0.500 & \bf0.885  & \bf0.667 \\  
            Meta-Llama-3.3    & 70.6 & \bf 0.444 & \bf 0.408 & $<$0.01  & $<$0.01 \\  
            Gemma-2-9B        & 9  & 0.386 & 0.371 & $<$0.01  & $<$0.01  \\  
            Gemma-3-1B        & 1  & 0.213 & 0.243 & \bf 0.029  &\bf 0.012  \\  
            Gemma-3-4B        & 4  & 0.376 & 0.364  & $<$0.01  & $<$0.01 \\  
            Gemma-3-27B       & 27 & 0.404 & 0.380 &$<$0.01  & $<$0.01 \\  
            \hline 
        \end{tabular}
    \end{threeparttable}
\end{table}

\subsection{Statistical Analysis}
To better understand LLM performance beyond overall correlations, we examined the distribution of predicted Ryff scores. Table~\ref{tab:stats} reports descriptive statistics (mean, median, standard deviation, minimum, and maximum) for all models alongside the ground-truth scores. While correlation metrics capture general predictive accuracy, they do not reveal the nature of prediction errors, such as systematic overestimation, underestimation, or limited variability. Examining the range, minimum, and maximum values helps assess whether predicted scores reflect the diversity of the ground-truth data or are instead constrained within a narrower interval.

\begin{table}[t]
    \centering
    \begin{threeparttable}
        \caption{Statistics of Ryff score predictions via LLMs}
        \label{tab:stats}
        \small
        \begin{tabular}{l|ccccc}
            \hline 
            Variable          & Mean & Median  & STD & Min & Max \\ 
            \hline
            Ground-Truth      & 91.7 & 95.0 & 16.8 & 51 & 123 \\
            \hline
            Meta-Llama-3.1    & 60.2 & 60.0 & 18.0 & 27 & 99 \\ 
            Ministral         & 63.8 & 64.0 &  9.8 & 35 & 84 \\ 
            Mistral-NeMo      & 80.7 & 80.0 &  9.6 & 56 & 100 \\  
            Microsoft-Phi-4   & 71.4 & 72.0 &  9.4 & 52 & 94 \\  
            QwQ-Preview       & 86.2 & 92.0 & 13.8 & 18 & 109 \\  
            DeepSeek-Qwen     & 79.3 & 83.5 & 25.1 & 0  & 126  \\  
            DeepSeek-Llama    & 82.0 & 90.0 & 21.2 & 53 & 103  \\  
            Meta-Llama-3.3    & 78.1 & 80.0 & 16.4 & 35 & 109 \\  
            Gemma-2-9B        & 67.0 & 69.0 & 12.9 & 27 & 92 \\ 
            Gemma-3-1B        & 61.9 & 64.0 & 11.2 & 29 & 93 \\  
            Gemma-3-4B        & 53.7 & 56.5 & 20.3 & 12 & 92 \\  
            Gemma-3-27B       & 60.4 & 62.0 & 14.9 & 26 & 93 \\  
            \hline 
        \end{tabular}
    \end{threeparttable}
\end{table}

The statistics in Table~\ref{tab:stats}, computed over all 111 participants, provide insight into prediction variability and error patterns. A dataset of this size, with spontaneous speech paired with clinically validated psychological questionnaires, is both rare and valuable, given the challenges of ethical approval, participant recruitment, and privacy. Statistically, it is sufficient to detect moderate-to-strong correlations at the standard significance threshold ($\alpha = 0.05$), ensuring that our analysis of LLMs' performance remains valid and meaningful.

As shown in Table~\ref{tab:stats}, most models tend to underestimate Ryff scores, though the degree varies. For instance, QwQ (median 92) is close to the ground truth (median 95), while Llama-3.1 has a much lower median of 60. Variability also differs across models: the ground-truth standard deviation is $\sim$17, with DeepSeek-Qwen overestimating at 25.1, Phi-4 underestimating at 9.4, and Meta-Llama-3.3-70B closely matching the ground truth at 16.4. Interestingly, DeepSeek-Llama, despite its statistically insignificant correlations, produces seemingly reasonable descriptive statistics: mean 82, median 90, and range 53--103, compared to ground-truth values of 91.7, 95, and 51--123. These metrics alone, however, do not capture the lack of statistical significance for this LLM.

\begin{figure}[t]
  \centering
  \includegraphics[width=\linewidth,height=140mm]{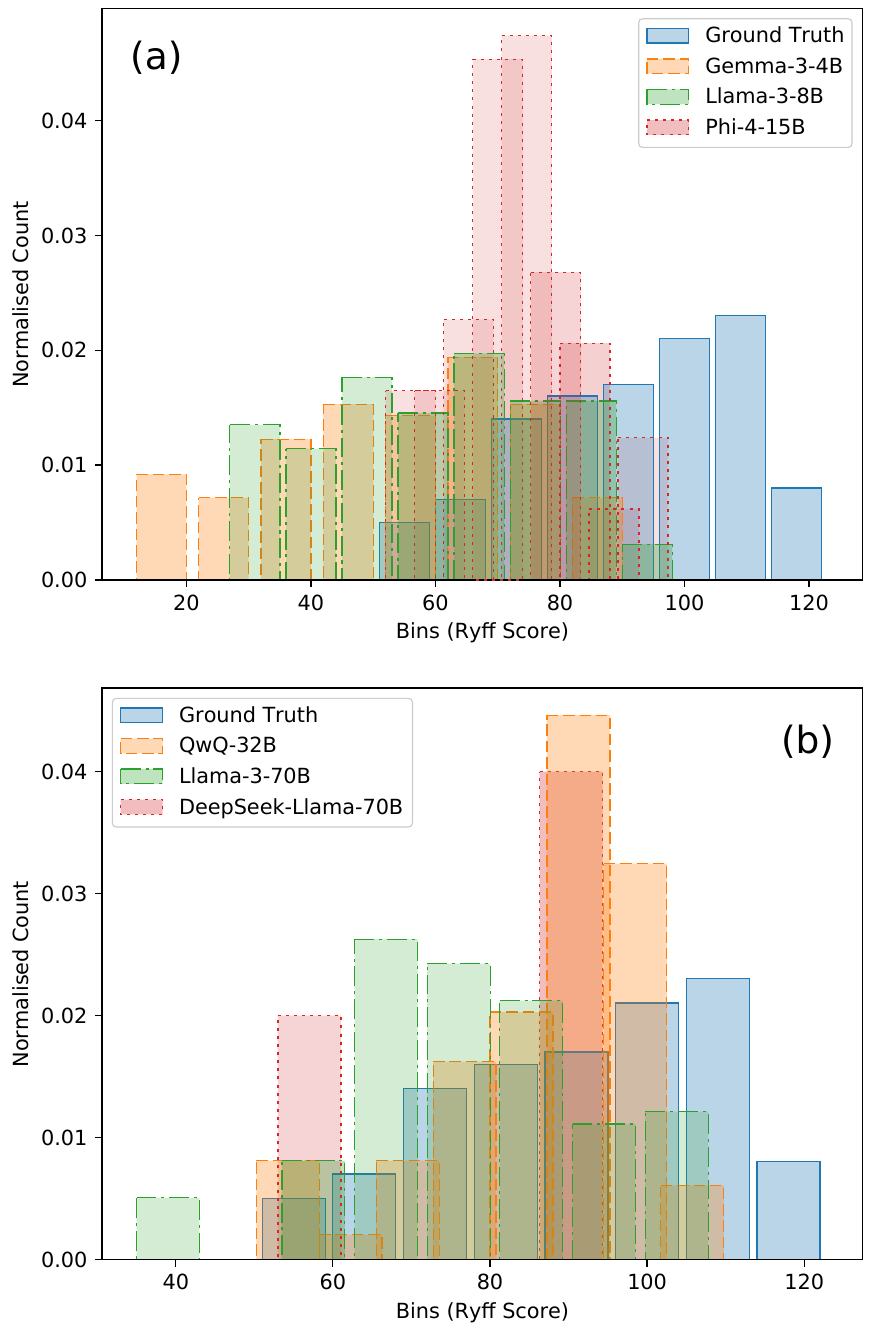}
  \caption{Histogram of the predicted Ryff scores by various LLMs vs ground truth.}
  \label{fig:hist-ryff}
\end{figure}

Fig.~\ref{fig:hist-ryff} shows histograms of predicted Ryff scores. DeepSeek\allowbreak-Llama exhibits a bimodal distribution with small standard deviations around each mode, explaining why its mean and median appear reasonable despite an insignificant correlation. Overall, most LLMs underestimate the Ryff scores and show non-bell-shaped distributions, highlighting the limitations of using Pearson correlation, which assumes Gaussianity.

\begin{figure}[h!]
  \centering
  \includegraphics[width=\linewidth, height=50mm]{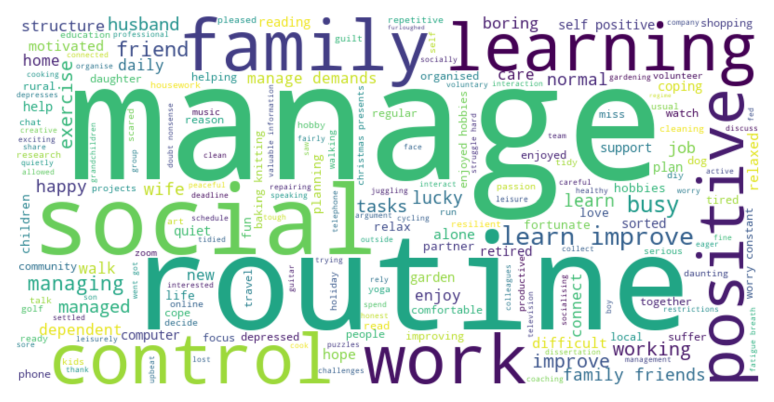}
  \caption{Word cloud of keywords extracted by Meta-Llama-3.3.}
  \label{fig:wordcloud}
\end{figure}

\subsection{Keyword Analysis via Word Clouds}
Interpretability is essential in healthcare applications, as performance metrics alone are insufficient for practical adoption. Word clouds, generated from keywords identified by the LLMs, provide an intuitive visualisation of the cues influencing predictions. To ensure reliability, we applied post-filtering to retain only words actually present in the transcripts. This step ensures that the highlighted terms genuinely reflect participants’ speech rather than model hallucinations, thereby enhancing the trustworthiness of the analysis.

As seen in Fig.~\ref{fig:wordcloud}, the most salient keywords are manage, family, learning, routine, social, control, and positive. These terms map closely onto Ryff’s six dimensions of psychological well-being. For example, \textit{manage} and \textit{control} directly relate to environmental mastery, reflecting an individual’s ability to handle daily demands and maintain agency. \textit{Family} and \textit{social} connect to positive relations with others, highlighting the importance of interpersonal support networks. \textit{Learning} signals personal growth, consistent with Ryff’s emphasis on self-development and openness to new experiences. \textit{Routine} reflects both autonomy and environmental mastery, as maintaining daily structure is a marker of coping during stressful contexts like lockdown. Finally, \textit{positive} aligns with self-acceptance and purpose in life, as expressions of optimism and self-evaluation often indicate higher well-being.

\section{Conclusion}
\label{sec:conclusion}


To our knowledge, this study is the first to demonstrate that LLMs, guided by a clinically grounded prompt, can estimate Ryff Psychological Well-Being (PWB) scores from spontaneous speech while providing interpretable justifications. Findings indicate that LLMs effectively extract psychologically meaningful cues, achieving Spearman correlations of up to 0.8 on information-rich transcripts. Statistical analysis revealed systematic underestimation and offered insight into prediction variability, while keyword-based word cloud analysis highlighted the specific linguistic features driving model decisions. 

Future research involves the multimodal fusion of acoustic and linguistic features to capture non-verbal affective signals overlooked by text, longitudinal trajectory analysis for the early detection of subtle mental health shifts, and the development of culturally adaptive prompts to address diverse expressions of well-being across global populations.

\section{Acknowledgment}
This work was supported by UKRI Grant No. 10102226 (University of Edinburgh) as part of the Horizon Europe Guarantee funding for participation in the INT-ACT project, supported by REA under the Horizon Europe Programme, Grant No. 101132719.
We would like to thank Angela Chitzanidi and the Research Services team at the University of Edinburgh for facilitating access to the Eddie Research Computing Cluster.

\bibliographystyle{IEEEbib}
\bibliography{refs}

\end{document}